\documentclass[runningheads]{llncs}
\usepackage{graphicx}
\usepackage{enumitem}
\usepackage{subfigure}
\usepackage[misc]{ifsym}

\begin{document}
\title{Responsibility Management through Responsibility Networks}
%
%

\author{Ruijun Chen \inst{1}, Jiong Qiu\inst{2} (\Letter)\and Xuejiao Tang \inst{3}
}


\institute{National Cheng Kung University\\
\email{n78083016@mail.ncku.edu.tw}\\
\and Hangzhou Quanshi Software Co., Ltd\\
\email{colin\_qiu@hotmail.com}
\and Leibniz University Hannover\\
\email{xuejiao.tang@stud.uni-hannover.de}
}

\maketitle              
%
\begin{abstract}
The safety management is critically important in the workplace. Unfortunately, responsibility issues therein such as inefficient supervision, poor evaluation and inadequate perception have not been properly addressed. To this end, in this demo paper, we introduce the Internet of Responsibilities (IoR) for responsibility management. Through the building of IoR framework, hierarchical responsibility management, automated responsibility evaluation at all level and efficient responsibility perception are achieved. The practical deployment of IoR system showed its effective responsibility management capability in various workplaces.

\end{abstract}
\section{Introduction}
Safety management places great demands on the supervisory ability and accountability in production environment~\cite{kang2020electronic,garcia2019selecting}. 
However, it is challenging to supervise and address responsibility due to the following reasons: 1) Difficult to supervise the main responsibility effectively. Since the professional talents are limited in a service company, it is difficult for the supervision to carry out every term and to realize selective examination regularly~\cite{malgieri2020concept}. Besides, if the supervision would be achieved by manual instead, more approaches are available to detect the potential safety hazard and supervise the main responsibility in detail. However, manual supervision are time-consuming and laborious, in which the potential safety hazard needs to be found periodically as well as checking whether the potential safety hazards have been eliminated by each enterprise on time~\cite{flyverbom2019governance}. 
2) Difficult to assess the post responsibility. Each station of enterprise is assigned with their own responsibility~\cite{chen2009corporate}. For example, the legal responsibility of enterprise is assigned by the post responsibility during the safety production. However, it is difficult to evaluate the responsibility performance of each station in this way. The personnel manager needs to detect which station is irresponsible and unqualified in the enterprise. Thus, the evaluation grade is important to reflect the performance of post responsibility. At present, signing the responsibility letter is the main way to post responsibility, which is difficult to carry out management quantitatively and effectively~\cite{tang2019internet}. The enterprises are therefore lack of efficient responsibility assessment.
3) The perception of information responsibility in IoT is not efficient. Nowadays, the assessment of post responsibility is improbable to realize by the artificial way scientifically. The difficulty is how to use the IoT~\cite{hosoda2018management}, such as the host machine of automatic fire alarm in the fire control system, to get information on a certain responsibility from the relevant station and assess the performance of their certain task, which could provide a better basis for personnel supervision~\cite{liu2019intelligent,baldo2013corporate}.

Therefore, it is more beneficial to manage and supervise the responsibilities in an automatic way. Our data driven IoR system serves this needs. Specifically, it allocates responsibilities for personnel and details the status of responsibility completion. In addition, it reminds the staff to fulfill responsibilities once the system detects a low responsibility score. The proposed systems have been deployed in various enterprises showing significant effectiveness in fulfilling different safety management requirements.  

\section{Responsibility Networks}
\begin{figure}[!htbp]
    \centerline{\includegraphics[width =0.9 \linewidth,height=4.7cm]{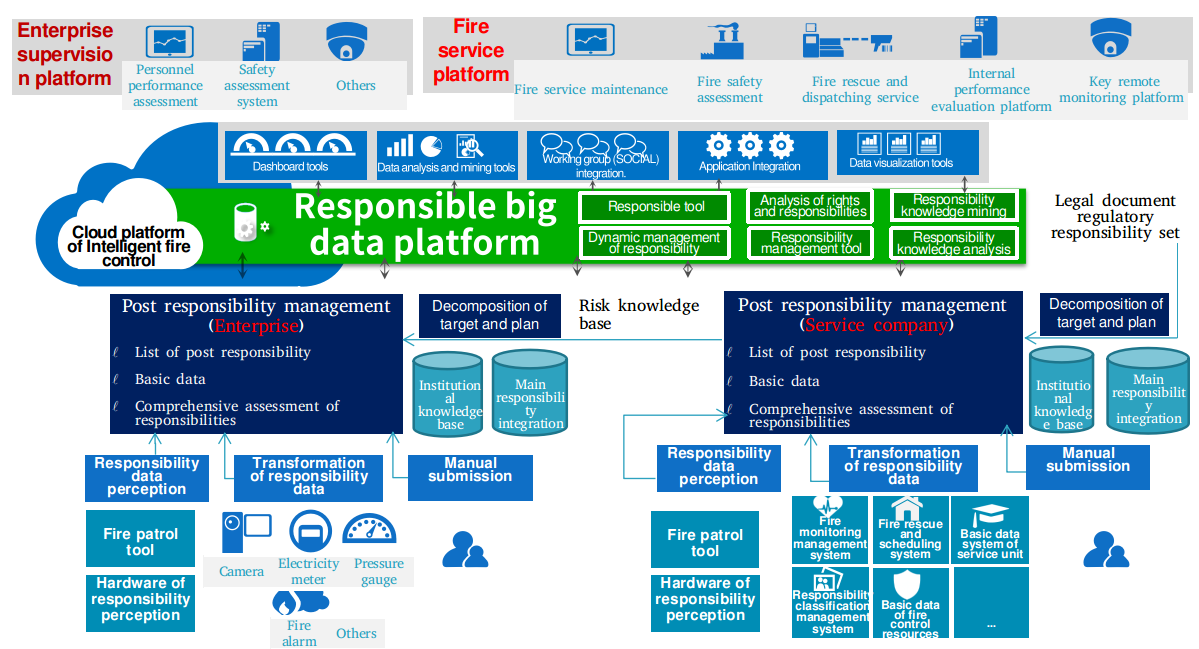}}
    \caption{The architecture of IoR system.}
    \label{fig:interface}
    \end{figure}

Our approach stems from the common idea of safety management. As shown in Figure~\ref{fig:interface}, the service company sorts out the responsibilities of fire protection laws, which could subdivide the corresponding targets for each station internally and the main responsibilities in the enterprise are supervised by a certain fire control supervisor. 

First of all, when the supervisor evaluates the detailed responsibility list, the database in system can be obtained from three ways: 1) perception of responsibility data; 2) IoT applications, such as fire supervision and management system, fire rescue scheduling system, enterprise’s basic data management system, responsibility classification management system, basic data of fire extinguishing resources, etc.; 3) manual input. Second, the integration of main responsibilities are delivered to each enterprise by service company bases on database and their characteristics, such as giving different requirements to huge shopping malls, hazardous chemical enterprises, etc. by means of fire control regulations. Then, the service company helps to combine the plan and objective of enterprise, which generates the responsibility list for each station according to the existing database of risk and system management in the enterprise. Next, each list needs to be weighted, which stands for the certain task of enterprises. The periodic task could also be set in each item, and the corresponding rules need to be established (cycle time, score method, etc.).
The big data of responsibility for each station will display and evaluate comprehensively after the enterprise executed duties. At the same time, the internal management system of the enterprise and the comprehensive management system of the service company are also combined in this IoT system. Thus, once the ``pressure'' in IoT system is higher than a certain critical value, the score of the corresponding responsibility list will be deducted automatically. The score can be deducted by days or by other criteria. The daily performance will be recorded once the station has executed duty. The further details of performance points in each month can be shown one by one, such as increasing and decreasing reasons.
\begin{figure}[!htb]
    \centering
    \includegraphics[width=0.9\linewidth,height=4.0cm]{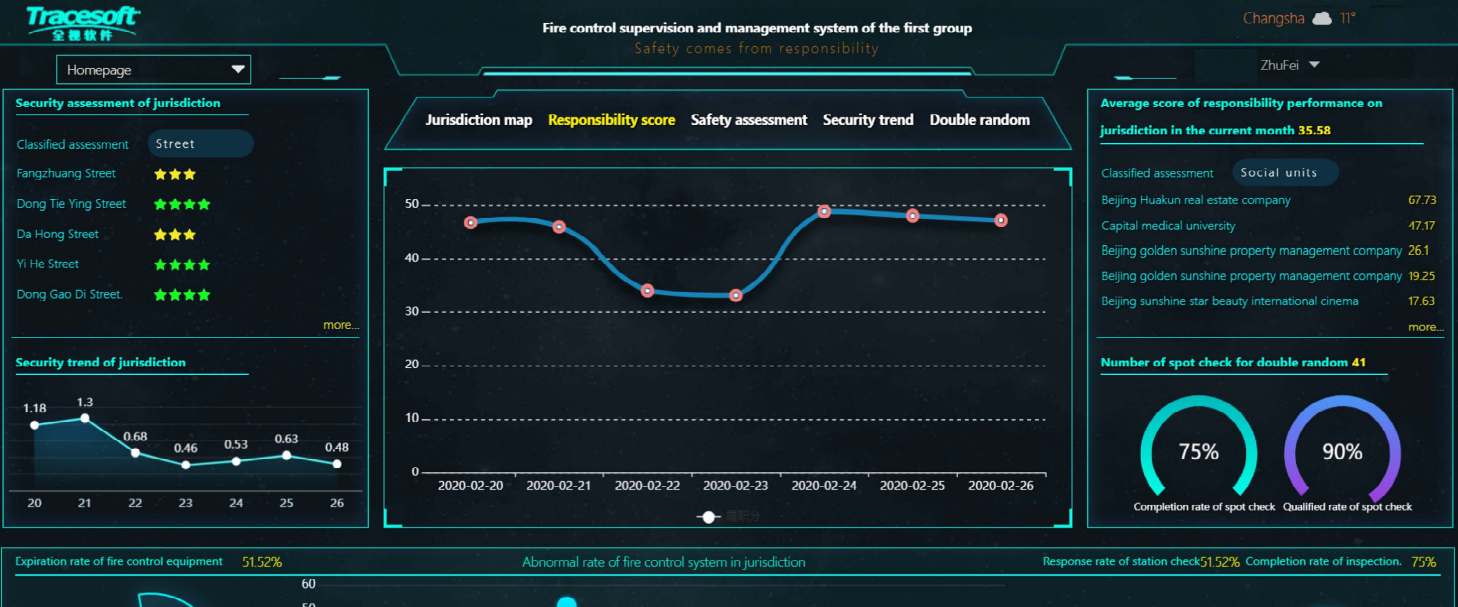}
    \caption{Exemplary graphical user interface of the IoR system}
    \label{fig:GUI}
\end{figure}
\begin{figure}[!htbp]
    \centering
    \subfigure[Responsibility score]{
    \includegraphics[width = 3.8cm]{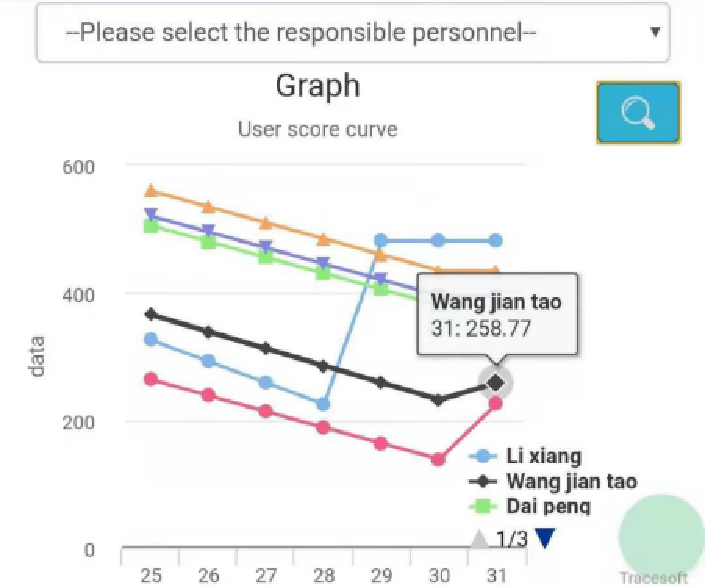}
    }
    \subfigure[Daily inventory data]{
    \includegraphics[width = 3.2cm]{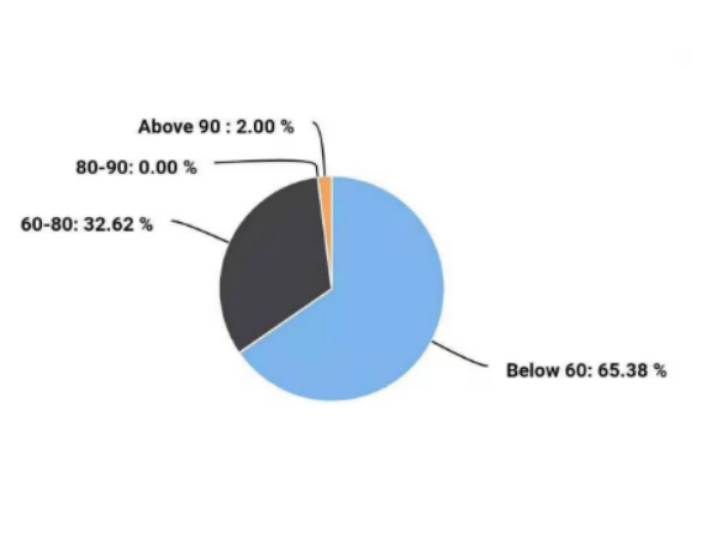}
    }
    \subfigure[Safety assessment results.]{
    \includegraphics[width = 4.3cm]{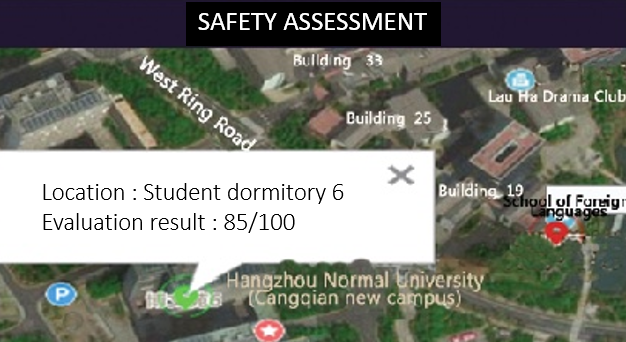}
    \label{fig:sar}
    }
  
\caption{The status of personnel responsibility completion.}
\label{fig: thresholded binary image}
\end{figure}
\section{System Demonstration}
The graphical user interface of the IoR system\footnote{The system is available at \url{www.ucgps.com/qsrespon}} is shown in Figure~\ref{fig:GUI}, which displays the average performance score and view the daily change. Different curves of performance in each station are shown in this figure. If an enterprise implements the main responsibility poorly, the system will find out which responsibilities are not in place and provide accountability to the enterprise leader and urge them to solve problems. Figure~\ref{fig: thresholded binary image} then lists the daily changes of personal responsibility scores, which could give each personnel a clear understanding of their own responsibility fulfillment. In addition, the safety evaluation results are shown in Figure~\ref{fig:sar}. The green buildings refer to high safety score regions while the buildings in low safety regions have color in red.

\section{Conclusion}
This paper introduces the practice of IoR management, which could help the service company to overview the performance of the main responsibility in each enterprise clearly and to achieve strong supervision. In IoR, practitioners can read through the responsibility score directly to check whether the responsibility of each station in the enterprise is in place, which is convenient and effective for managing the station, such as accountability and reward. The evaluation score for the station can also interchange with the information of IoT for a quick and scientific responsibility assessment.

\nocite{*}
\bibliographystyle{IEEEtran}
\bibliography{ref}

\end{document}